\documentclass{article}
\usepackage{ijcai24}

\usepackage{adjustbox}
\usepackage{times}
\usepackage{soul}
\usepackage{url}
\usepackage[hidelinks]{hyperref}
\usepackage[utf8]{inputenc}
\usepackage[small]{caption}
\usepackage{graphicx}
\usepackage{amsmath}
\usepackage{amsthm}
\usepackage{booktabs}
\usepackage{algorithm}
\usepackage{algorithmic}
\usepackage[switch]{lineno}
\usepackage{booktabs}
\usepackage{bibentry}
\usepackage{ragged2e} 
\usepackage{booktabs,makecell, multirow, tabularx}
\usepackage{amssymb}

\usepackage{colortbl}
\usepackage{amsmath,amssymb,amsfonts}
\usepackage{algorithmic}
\usepackage{graphicx}
\usepackage{float}

\title{Delve into Base-Novel Confusion: Redundancy Exploration for Few-Shot Class-Incremental Learning}

\author{
Haichen Zhou$^1$
\and
Yixiong Zou$^{1}$\thanks{denotes the corresponding author.}\and
Ruixuan Li$^{1}$\and
Yuhua Li$^1$\And
Kui Xiao$^2$\\
\affiliations
$^1$Huazhong University of Science and Technology\\
$^2$Hubei University\\
\emails
\{m202273832, yixiongz, rxli, idcliyuhua\}@hust.edu.cn, xiaokui@hubu.edu.cn}

\begin{document}
\maketitle

\begin{abstract}
	Few-shot class-incremental learning (FSCIL) aims to acquire knowledge from novel classes with limited samples while retaining information about base classes. 
    Existing methods address catastrophic forgetting and overfitting by freezing the feature extractor during novel-class learning. 
    However, these methods usually tend to cause the confusion between base and novel classes, i.e., classifying novel-class samples into base classes.
    In this paper, we delve into this phenomenon to study its cause and solution.
    We first interpret the confusion as the collision between the novel-class and the base-class region in the feature space.
    Then, we find the collision is caused by the label-irrelevant redundancies within the base-class feature and pixel space. 
    Through qualitative and quantitative experiments, we identify this redundancy as the shortcut in the base-class training, which can be decoupled to alleviate the collision.
    Based on this analysis, to alleviate the collision between base and novel classes, we propose a method for FSCIL named Redundancy Decoupling and Integration (RDI).
    RDI first decouples redundancies from base-class space to shrink the intra-base-class feature space. 
    Then, it integrates the redundancies as a dummy class to enlarge the inter-base-class feature space. 
    This process effectively compresses the base-class feature space, creating buffer space for novel classes and alleviating the model's confusion between the base and novel classes.
    Extensive experiments across benchmark datasets, including CIFAR-100, \textit{mini}ImageNet, and CUB-200-2011 demonstrate that our method achieves state-of-the-art performance.
\end{abstract} 
\section{Introduction}
\begin{figure}[t!]
	\centering
	\includegraphics[width=7cm]{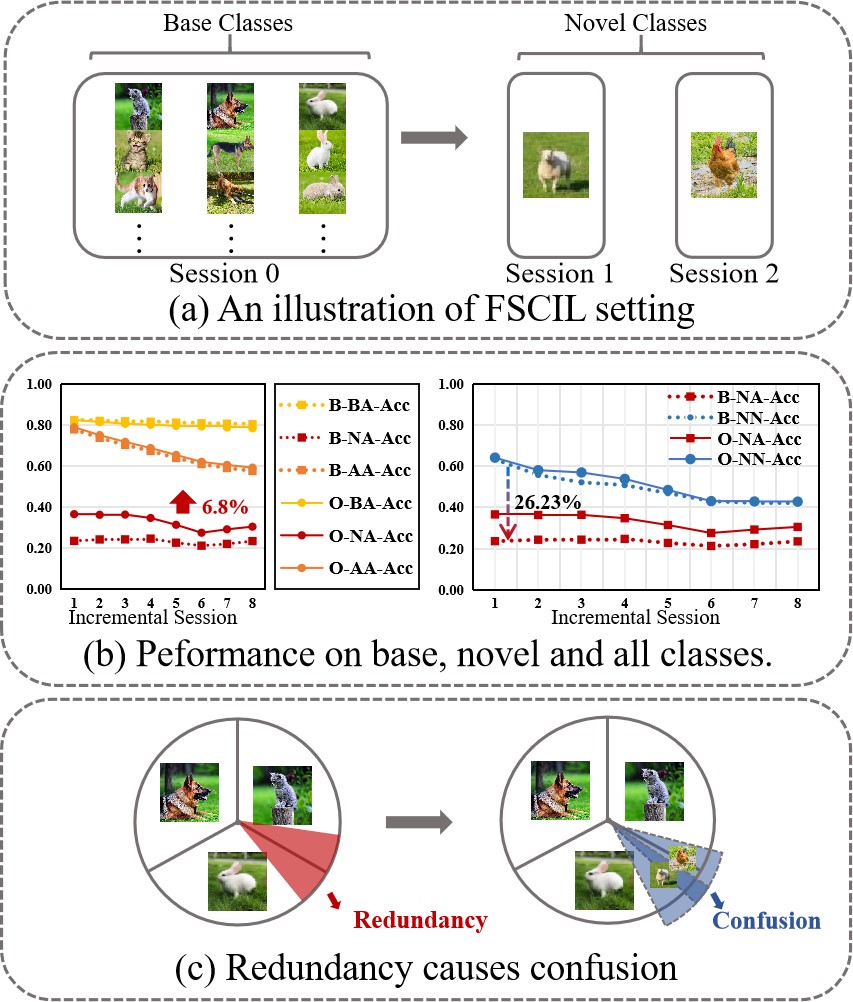}
	\caption{(a) FSCIL entails mastering base classes through ample examples and incrementally addressing novel classes with few samples. (b)  Left: Performance on base classes, novel classes, and all classes. BA-Acc, NA-Acc, and AA-Acc denote the average accuracy of classifying Base-class, Novel-class, and All-class samples against All encountered classes, respectively. B- represents using the Baseline fixed-backbone method, while O- represents using Ours. Right: Performance on novel-class samples. NN-Acc denotes the average accuracy of classifying Novel-class samples against only encountered Novel classes. We can see the model tends to classify novel-class samples into base classes, which is a major cause of the low performance in FSCIL. We term this problem as the confusion between novel and base classes, which we aim to interpret and alleviate in this paper. (c) Intuitive interpretation of the confusion in the feature space and the pixel space.
}\label{f1}
\end{figure}

Deep Neural Networks have achieved remarkable success in many domains \cite{liu2020weakly,li2022rigidflow,daihong2022multi}. However, real-life scenarios often involve streaming data, posing challenges to obtaining sufficient data for continuous learning of new knowledge, especially with limited samples.
To address this, Few-Shot Class-Incremental Learning (FSCIL) \cite{tao2020few} was introduced. FSCIL entails learning base classes with an adequate number of samples in the base session, followed by learning novel classes using only a few samples in incremental sessions (Fig.\ref{f1}a).

The primary challenges in Few-Shot Class-Incremental Learning (FSCIL) center around the catastrophic forgetting of learned knowledge and the overfitting \cite{hu2019overcoming} to novel classes due to limited data. To address these challenges, current approaches typically initiate training on base classes and then freeze the backbone while learning novel classes \cite{snell2017prototypical}. However, although the fixed-backbone method maintains the model's performance on base classes, it impedes effective generalization to novel classes. 

To investigate this issue, we briefly plot the performance of classifying base-class, novel-class, and all-class samples into all encountered classes, denoted as BA-Acc, NA-Acc, and AA-Acc, separately in Fig.~\ref{f1}b(left). We can see the yellow dashed line notably surpasses the red dashed line, signifying a considerable decrease in performance on novel-class samples compared to base-class ones.
Then, we divide the novel-class performance into classifying novel-class samples against only novel classes and against all (i.e., base + novel) classes, denoted as NN-Acc and NA-Acc, respectively in Fig.~\ref{f1}b(right). 
We can observe the prediction accuracy significantly drops by 26.23\% when base classes are introduced, which is a major cause of the low performance on novel-class samples. This phenomenon indicates the model tends to misclassify novel-class samples into base classes, and we call it the confusion between base and novel classes.
Although this phenomenon widely exists in current FSCIL methods, only a few works \cite{wang2023few} attempted to explore its cause.

In this study, we delve into the cause of the confusion between base and novel classes, focusing on the redundancy in the feature space and the pixel space. We interpret the confusion as the collision between the novel-class region and the base-class region in the feature space, which is caused by the redundant area of base-class regions (Fig.~\ref{f1}c). 
Through qualitative and quantitative experiments, we find such redundancy reflects shortcuts of the model training, which can be decoupled to alleviate the collision between base and novel classes.

Based on the above analysis, to alleviate the confusion between base and novel classes, we propose an approach named Redundancy Decoupling and Integration (RDI) to address the redundancy within the base-class feature space. 
We first decouple the redundancy in the feature space by finding the label-irrelevant regions, and then map these regions to the pixel space for image masking, which would shrink the intra-base-class space. Then, we integrate these label-irrelevant regions by classifying masked images into a dummy class, which further compresses the base-class feature space and reserves a buffer space for novel classes. Consequently, the reserved space would reduce the collision between base and novel-class regions in the feature space, therefore alleviating the confusion between base and novel classes.

The contributions of this work are summarized as follows:
\begin{itemize}
	\item Our experimental analysis reveals that the model's generalization on novel classes is majorly limited by the confusion between base and novel classes, predominantly attributed to label-irrelevant redundancies in the base-class feature space.
         \item Based on the analysis, we propose an innovative approach to mitigate the confusion between novel and base classes, which firstly decouples the label-irrelevant redundancies in both the feature space and the pixel space, and then integrates these redundancies to reserve a butter space for novel classes.
	\item Experimental results not only showcase the efficacy of our approach in enhancing generalization capacity on novel classes but also demonstrate its achievement of state-of-the-art performance on benchmark datasets, including CIFAR-100, CUB200, and \textit{mini}ImageNet.
\end{itemize}

\section{Interpretation of Base-Novel Confusion}
In this section, we initially outline the Few-Shot Class-Incremental Learning (FSCIL) task and the baseline method. 
Then, we conduct experiments to interpret the cause of the confusion between base and novel classes.

\subsection{Problem Set-up}
\textbf{Few-shot Class Incremental Learning (FSCIL)} is a learning paradigm that involves mastering base classes with ample examples before tackling novel classes with limited samples. Typically, FSCIL employs a sequence of $T$ training sessions denoted as $\smash{\{D_{\text{train}}^{0},\ldots,D_{\text{train}}^{T-1}\}}$, each associated with a distinct label space $\smash{\{C^{0},\ldots,C^{T-1}\}}$, where there is no overlapping between label spaces. 

During each session, only the session-specific training set $\smash{D_{\text{train}}^{t}}$ is available, comprising examples $\smash{\{(x_{i},y_{i})\}_{i=1}^{n^{t}}}$ with $\smash{y_{i} \in C^{t}}$, and $\smash{n^{t}}$ denotes the number of examples. The testing set $\smash{D_{\text{test}}^{t}}$, however, includes samples from all previously encountered and current classes, denoted as $\smash{D_{\text{test}}^{t}=\{(x_{i},y_{i})\}_{i=1}^{m^{t}}}$, where $\smash{y_{i} \in \bigcup_{i=0}^{t}C^{i}}$. The initial session $(t=0)$ involves a training set with ample examples per class. The model is built as $\smash{f(x) = W^{T}\phi(x)}$, where $\smash{\phi(x)}$ represents feature extraction and $\smash{W}$ denotes the classifier for base class categorization. Subsequent incremental sessions $(0 < t < T)$ consisting of limited samples are organized as N-way K-shot tasks. In these sessions, the model learns novel classes while maintaining performance on base classes.

\subsection{Baseline Description}
The baseline model follows a structured approach for incremental learning. In the base session, the model undergoes training on $D_{train}^{0}$ utilizing the cross-entropy loss:
\begin{equation}\label{soft}
L=\frac{1}{N}  \sum_{i=1}^{N} -log \frac{e^{\tau cos(x_{i},w_{c} )} }{e^{\tau cos(x_{i},w_{c} )}+ {\textstyle \sum_{m\neq c}^{}e^{\tau cos(x_{i},w_{m} )}} } 
\end{equation}

Moving to the incremental session, each base class's classifier is replaced with the class's average embedding, known as the prototype. The prototype calculation is denoted as 
\begin{equation}\label{proto}
p_c = \frac{1}{n^0_c} \textstyle \sum_{j=1}^{D_{train}^0} \mathbb{I}(y_j=c) \phi (x_j)
\end{equation}
where $\mathbb{I}$ represents the indicator function, $\phi (\cdot)$ is the feature extractor, and $n^0_c$ denotes the count of examples in class $c$. 
During the incremental session, the classifier for base classes remains fixed, while for novel classes, the classifier can be refined through fine-tuning or by employing the average embedding. During evaluation in session $T$, an input $x$ is predicted using its output feature $\phi (x)$ and the classifier $W$, by: 
\begin{equation}\label{infer}
\hat{y}=argmax_{k} \quad cos(\phi (x),w_k), \forall 1\le k\le \sum_{i=0}^{T} N^i
\end{equation} 

\subsection{Interpretation of Confusion Between Novel and Base Classes}
In Fig.~\ref{f1}b, we conclude that the major limitation of the novel-class generalization: the model tends to classify novel-class test samples into base classes, termed "Base-Novel Confusion". In this section, we delve into this confusion from the redundancy in the feature space.
\begin{figure}[htbp]
        \vspace{-0.2cm}
	\centering
	\includegraphics[width=7.5cm]{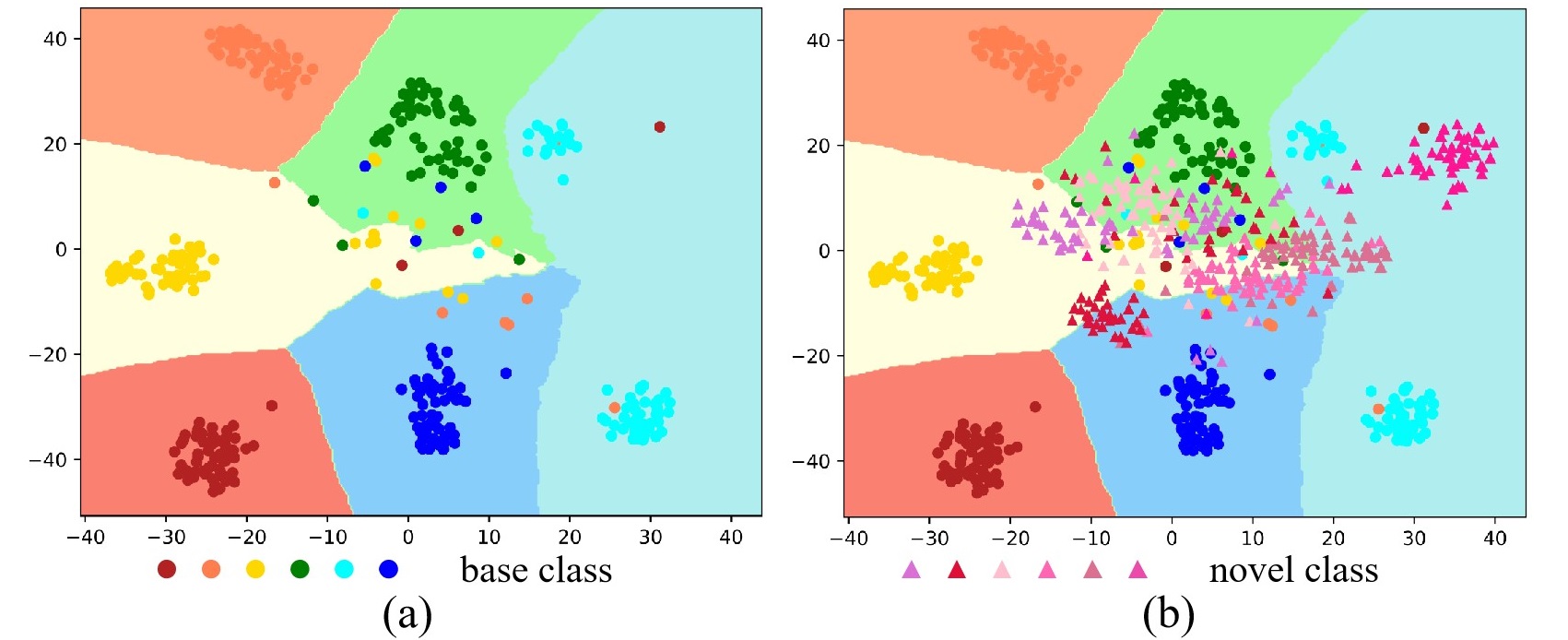}
        \vspace{-0.2cm}
	\caption{(a) Feature distribution and decision boundary of six base classes. (b) Feature distribution of six introduced novel classes, based on the six existing base classes. The introduced novel classes demonstrate a potential for confusion with the base classes.}
	\label{m01}
        \vspace{-0.4cm}
\end{figure}
\subsubsection{Confusion represented as collisions in the feature space.}
To explore the confusion between base and novel classes, we first plot the feature distribution for both base and novel classes in Fig.~\ref{m01}. Initially, we randomly select sixty samples from each of six randomly sampled base classes in CIFAR-100 and visualize the distribution of feature points (circles) and decision boundaries using t-SNE \cite{van2008visualizing}, as shown in Fig. \ref{m01}a. 
To simulate the baseline model's learning process of novel classes, we then visualize the feature distribution of six newly introduced (randomly sampled) novel classes together with base classes, as depicted in Fig. \ref{m01}b. Notably, these novel classes, represented as triangles, fall within the space allocated for base classes, with some even closely approaching the feature points (circles) representing base classes. 
Although we can see clear clusters even for novel classes, novel samples are located within the decision boundary for each base class, leading to collisions in feature spaces and confusing the model to classify novel-class samples into base classes.
In essence, the collision exists because the model learns to occupy the entire feature space during the base-class training.
Therefore, a natural question arises: \textit{do base classes really need to occupy so much space to make the model discriminative on them?}
\begin{figure}[htbp]
	\centering
	\includegraphics[width=6cm]{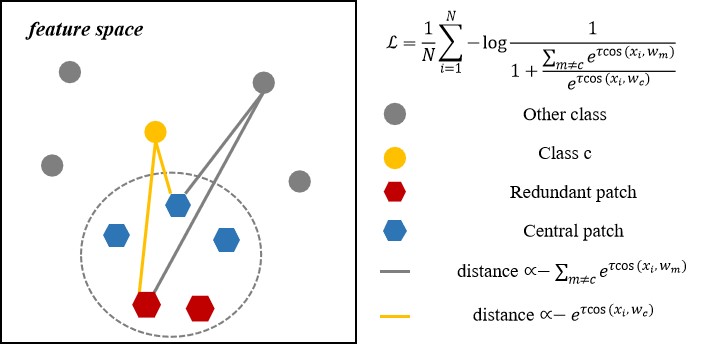}
	\caption{Because of the misalignment between decreasing $\sum_{m\neq c}^{}e^{\tau cos(x_{i},w_{m} )}$ and increasing $e^{\tau cos(x_{i},w_{c} )}$, the ordinary cross-entropy loss could push the model to learn redundant features that are both distant from other classes and the class $m$.}
	\label{loss}
 \vspace{-0.3cm}
\end{figure}
\subsubsection{Redundancies caused by the ordinary classification loss.}
To answer the above question, we look back into the classification loss utilized to train the baseline model. The ordinary cross-entropy loss in Eq.\ref{soft} can be rewritten as illustrated in Fig.\ref{loss}. To minimize this loss, the model needs to increase $e^{\tau cos(x_{i},w_{c} )}$ and decrease $\sum_{m\neq c}^{}e^{\tau cos(x_{i},w_{m} )}$. In other words, the model learns to (1) push features in the class $c$ close to class $c$'s centroid while (2) pulling away these features from other classes. 
However, these two targets may not be aligned, because \textbf{features away from other classes may not be close to the class $c$}, which could lead to redundancies in the feature space.
For example, in Fig.\ref{loss}, although red hexagons (could be features of an instance or an image patch) are pulled away from other classes (the gray circle) by decreasing $\sum_{m\neq c}^{}e^{\tau cos(x_{i},w_{m} )}$, it may not be close to class $c$'s centroid (the yellow circle).
This means the features represented by red hexagons, although may help the classification of the class $c$, may not really capture the central characteristics of the class $c$, which could be understood as a kind of shortcut that overfits the class $c$.

\begin{table}[htbp]
	\centering
	\caption{Similarities between patches and the ground truth class as well as other classes. Redundant patches demonstrate smaller similarities from both kinds of classes, consistent with Fig.\ref{loss}.}
    \vspace{-0.2cm}
	\adjustbox{width=8.5cm}{
		\begin{tabular}{l|cc|cc}
			\toprule
			\multicolumn{1}{c}{\multirow{2}{*}{\textbf{Dataset}}} & \multicolumn{2}{c}{$ e^{\tau cos(x,w_{c} )} $} & \multicolumn{2}{c}{$\sum_{m\neq c}^{}e^{\tau cos(x,w_{m} )}$}\\ 
			& Redundant Patches & Central Patches& Redundant Patches & Central Patches  \\ \cmidrule(r){1-5}
			 CIFAR-100    & 4.50  & 69.69        & 1.70  & 3.10      \\
			\textit{mini}ImageNet   & 4.21 & 36.66      & 0.87   & 1.60  \\
			CUB200 &7.28   & 20.72    & 1.31   & 5.67          \\ \bottomrule
	\end{tabular}}
	\label{distance}
 \vspace{-0.2cm}
\end{table}

To validate our hypothesis, we compute the similarity of each patch $f(a,b) \in \mathbb{R}^{d}$ of the feature map $ F \in \mathbb{R}^{h \times w \times d}$ using a pre-trained model from its ground truth class, and categorize the patches into central and redundant patches by thresholding the similarity. 
The central patch is close to its ground truth class, while the redundant patch is far from that class. 
Subsequently, we utilize $e^{\tau cos(x_{i},w_{c} )}$ and $\sum_{m\neq c}^{}e^{\tau cos(x_{i},y_{m} )}$ to compute the average similarity between patches and classes, as depicted in Tab.\ref{distance}. 
Redundant patches are not only distant from both the ground truth class but also far from other classes. 
This quantitatively indicates that the model tends to learn label-irrelevant but discriminatory redundant patches to minimize $\sum_{m\neq c}^{}e^{\tau cos(x_{i},y_{m} )}$, thereby reducing the overall loss.
\vspace{0.1cm}
\subsubsection{Redundancies as label-irrelevant regions in the heatmap.}
\begin{figure}[htbp]
	\centering
        \vspace{-0.2cm}
	\includegraphics[width=7cm]{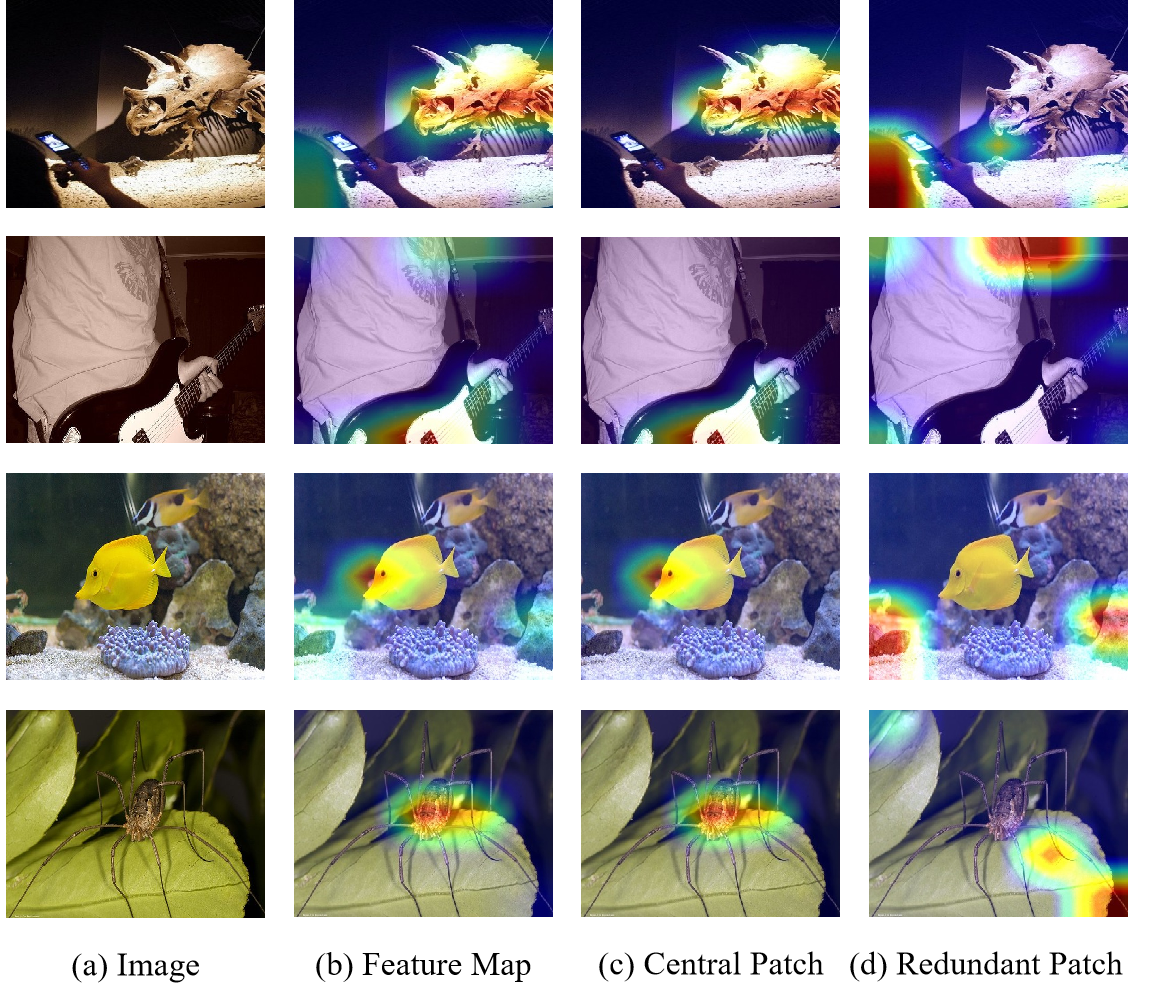}
        \vspace{-0.2cm}
	\caption{Decoupling of feature map into central and redundant patches. The redundant patches correspond to label-irrelevant objects in the base-class pixel space.}
	\label{v1}
 \vspace{-0.3cm}
\end{figure}

To ascertain whether redundancies are genuinely label-irrelevant, we examine them in the pixel space. Initially, we scrutinize the feature map of base classes under the pre-trained baseline model, visualizing the output from the last convolutional layer as a heat map using ProtoNet \cite{snell2017prototypical} (Fig.\ref{v1}, 'Feature Map'). Although the pre-trained model can identify semantic objects with higher activations, not all activated regions necessarily align with the label. For instance, in the visualization (third row of Fig.\ref{v1}), activated regions corresponding to both a fish and a part of stones were identified, despite the image being labeled as a fish. The parts of the stones are considered irrelevant to the label, while the fish is relevant. We subsequently isolate central and redundant patches from the original feature map by their similarities with the ground truth class following Tab.\ref{distance}. 
We can see the redundant patches correspond to label-irrelevant but discriminatory regions. Based on the above experiments, we conclude that the ordinary cross-entropy loss leads the model to focus not only on label-relevant features (central patches) but also on label-irrelevant redundancies for discrimination.

Since redundant features also occupy spaces within the decision boundary of each base class, 
it could lead to the collision between base and novel classes in the feature space. 
This inspires us to revisit the feature space to explore the relationship between redundancies and collision. 
\vspace{0.1cm}
\subsubsection{Redundancies lead to collisions in the feature space.} 
\begin{figure}[htbp]
	\centering
	\includegraphics[width=7cm]{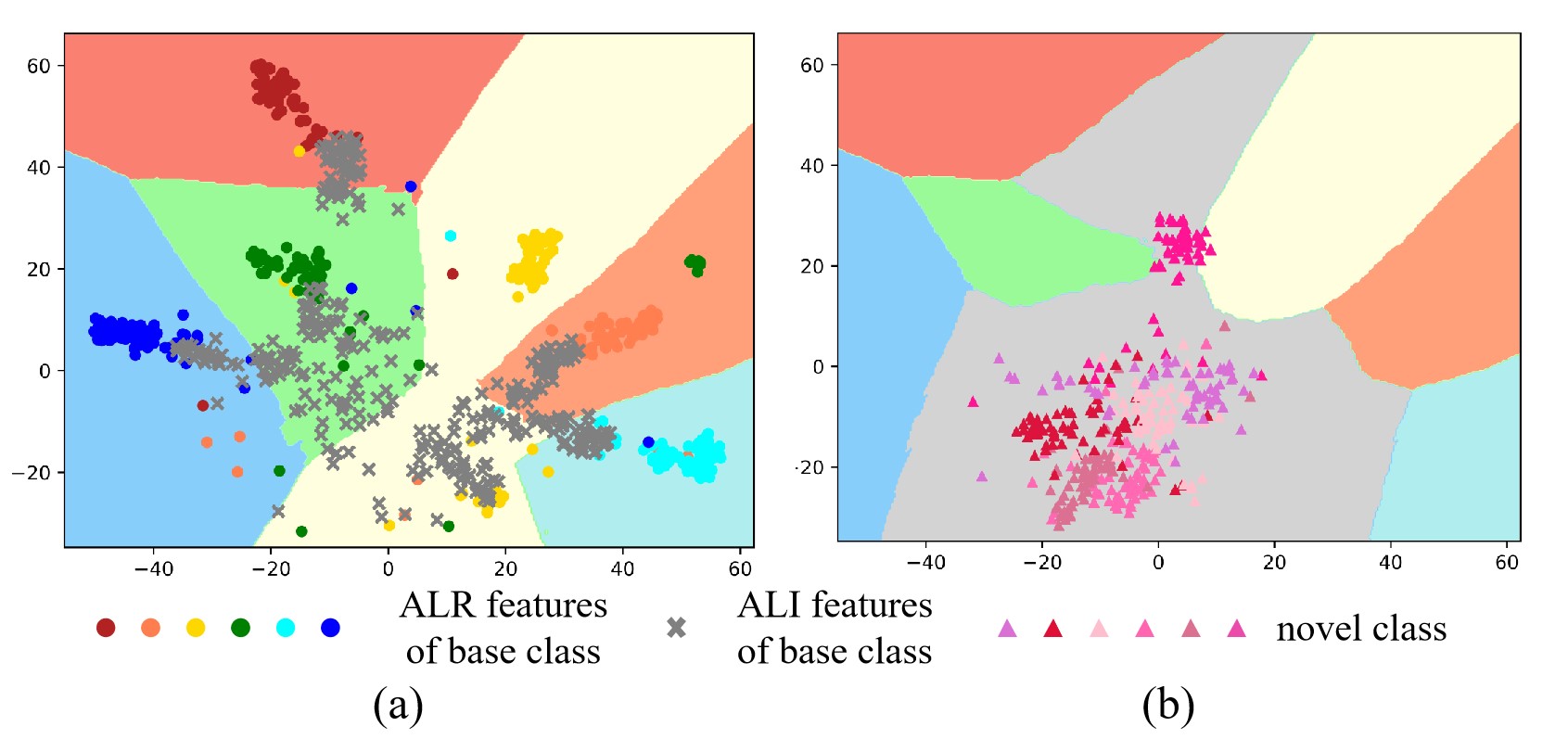}
         \vspace{-0.3cm}
	\caption{(a) Visualization of the distribution of ALR and ALI features from base classes. Pixel-space Redundancies (ALI features) imply redundancy in feature space and enlarge the base-class space. (b) Visualization of the distribution of newly introduced novel classes, based on the existing ALR and ALI features. The gray areas indicate regions of base-class redundancies. We can see novel classes are located in this gray region, indicating that redundancy in the base-class feature space leads to confusion.}
	\label{m02}
 \vspace{-0.3cm}
\end{figure}

We first decouple the Activated Label-Irrelevant (ALI) features and Activated Label-Relevant (ALR) features from the original feature map, where ALI corresponds to the redundant patches and ALR corresponds to the central patches. 
Then, we visualize the feature distribution and decision boundaries of ALI features and ALR features from the six previously selected classes, as shown in Fig.\ref{m02}. 
We can see that ALI features (gray fork) are distributed around ALR features, and ALR features exhibit a more compact distribution compared to the original feature distribution. In other words, these ALI features, acting as redundancies in feature space, expand the base-class space. Without these redundancies, the base-class space would be more compact.

Therefore, we hypothesize that these redundancies in the base-class feature space are the primary cause of the model's confusion between base and novel classes. To substantiate this, we integrate all ALI features as a dummy class and visualize the decision boundaries of base classes (using ALR features) and the dummy class (grey region). 
Then, we plot the feature distribution of novel classes (Fig.\ref{m02} right). 
We observe that novel classes are concentrated in the grey region where redundant ALI features are distributed, which confirms our speculation that the redundancies lead to the collision.

\vspace{0.1cm}
\subsubsection{Conclusion and Discussion.}
We conclude that the confusion between novel base classes can be interpreted as the collisions in the feature space. Based on the analysis of the regular classification loss, we find that it will encourage the model to learn label-irrelevant but discriminatory patches, which could be redundancies. To investigate the relationship between redundancies and collision in feature space, we visualize the redundancies in feature space and find that it is the redundancies in base-class space that mainly cause the collision between base and novel classes. 

\section{Proposed Method}
\begin{figure*}[t]
	\centering
	\includegraphics[height=3.5cm]{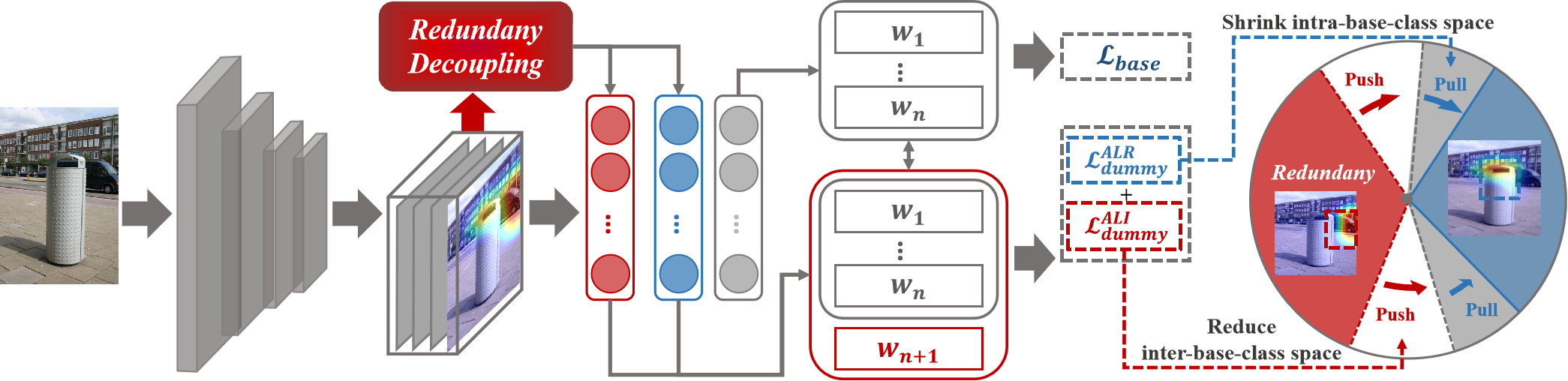}
	\caption{Our FSCIL framework: (1) Decoupling redundant ALI features from the base-class space and utilizing ALR features for base-class classification, minimizing intra-base-class space (loss term $\mathcal{L}^{ALR}_{dummy}$). (2) Integrating redundant ALI features to generate samples of a dummy class for ALR-based recognition, reserving space for novel classes and reducing inter-base-class space (loss term $\mathcal{L}^{ALI}_{dummy}$). }
	\label{frame}
 \vspace{-0.4cm}
\end{figure*}

While the baseline approach addresses catastrophic forgetting to some extent by employing a fixed feature extractor, it compromises the model's capacity to generalize over novel classes. This raises a natural question: 

\textbf{How can we improve novel-classes generalization while maintaining base-classes performance?}

Based on the above analysis, we pinpoint the key factors leading to inadequate generalization of existing methods in novel classes—the confusion between novel and base classes. This confusion predominantly stems from redundancy within the base-class feature space. In response, we introduce our method, Redundancy Decoupling and Integration (RDI), for Few-Shot Class Incremental Learning (FSCIL). RDI comprises two modules. The first module decouples redundancy—Activated Label-Irrelevant (ALI) features—from the base-class feature space, utilizing Activated Label-Relevant (ALR) features for base-class classification. In the second module, we integrate redundancies as a dummy class to contribute to ALR-based recognition. An overview of our method is illustrated in Fig.\ref{frame}.

\subsection{Shrink Intra-Base-Class Space by Decoupling Redundancy from Base-Class Space}

For a more precise decoupling of the base-class space, we propose a two-step approach. Firstly, we pre-train the model using the cross-entropy loss, obtaining a classifier $\smash{W_{\text{base}}=[w_1,\ldots,w_n] \in \mathbb{R}^{d\times n}}$ and a feature extractor $\smash{\Phi(\cdot)}$. Here, $n$ is the number of base classes. Next, we initiate the decoupling process by segmenting the feature map output from the pre-trained model into ALR (\textit{Activated Label-Relevant}) and ALI (\textit{Activated Label-Irrelevant}) regions.

Based on the analysis in Section 2, we aim to decouple redundancies by computing the distance between each patch $ f(a,b) \in \mathbb{R}^{d}$ of the feature map $ F \in \mathbb{R}^{h \times w \times d}$ and its ground truth class and generating an ALR mask $M^{ALR} \in \mathbb{R}^{h \times w} $:
\begin{equation}\label{eq4}
	M^{ALR}(a,b) = \sigma (\sum_{k=1}^{d} f_{k}(a,b)w_{y_{p}}^k)
\end{equation}
where $\sigma(\cdot)$ is an activation function with a threshold $\varrho $ as a hyperparameter, $y_p$ is the predicted label, and $M^{ALR}\in \mathbb{R}^{h\times w} $ is a 0-1 mask, Considering the possibility of misclassified samples and incorrectly activated feature maps after the pre-training stage, we suggest using the predicted label instead of the real label for all samples to ensure the validity of the feature map. The predicted label $y_p$ can be obtained by:
\begin{equation}\label{eq3}
	y_{p}=argmax{ \quad (\frac{W_{base}}{\left\|W_{base}\right \|_{2} })^T (\frac{\Phi (x)}{\left \|\Phi (x)\right\|_{2}})}
\end{equation}
Subsequently, the ALR feature $f^{ALR}$ is obtained as:
\begin{equation}\label{eq5}
	f^{ALR}=Pooling(M^{ALR}F)
\end{equation}
We label the ALR feature with its real label, i.e., $y^{ALR} = y$. Consequently, the cross-entropy loss of using the ALR feature for classification can be represented as:
\begin{equation}\label{eq6}
	\mathcal{L}^{ALR}_{base}=CE\left ((\frac{W_{base}}{\left\|W_{base}\right \|_{2} })^T (\frac{f^{ALR}}{\left \| f^{ALR}\right\|_{2}}),y \right ),
\end{equation}
The classification loss using the original feature is denoted as $\mathcal{L}_{base}$. The total loss is then expressed as a weighted sum of the original loss and the ALR loss: $\mathcal{L}= \mathcal{L}_{base} + \lambda \mathcal{L}^{ALR}_{base}$, where $\lambda$ is a hyperparameter determining the balance between the two losses.

Through $\mathcal{L}^{ALR}_{base}$, the model guides central patches to approach the ground truth class and distance themselves from other classes. This strategy mitigates shortcuts in the base-class training, effectively decoupling redundancies from the base-class space.
\subsection{Reduce Inter-Base-Class Space by Integrating Redundancies as a Dummy Class}
In contrast to previous methods, our innovative approach integrates redundancies as a dummy class to contribute to ALR-based recognition. This process compresses intra-base-class space and reserves a space for novel classes, effectively mitigating confusion between novel and base classes.

Building on the insights from Section 2, we categorize patches in the feature map into two groups: central patches (ALR) and redundant patches (ALI). Subsequently, we obtain the ALI mask from the ALR mask, denoted as $M^{ALI}(a,b)=1-M^{ALR}(a,b)$. Utilizing this ALI mask, we derive the ALI feature $f^{ALI}\in \mathbb{R}^{d}$.
\begin{equation}\label{eq7}
	f^{ALI}=Pooling(M^{ALI}F)
\end{equation}
Afterward, the ALI feature is labeled as $n+1$, indicated as $y^{ALI}=n+1$. To set aside space for novel classes, we adjust the classifier by incorporating the dummy class, yielding $W_{dummy} = [W_{base},w_{n+1}] \in \mathbb{R}^{d\times(n+1)}$. Subsequent to this modification, the loss for the ALR feature can be updated:
\begin{equation}\label{eq8}
	\mathcal{L}^{ALR}_{dummy}=CE\left ((\frac{\scriptstyle W_{dummy}}{\left\|\scriptstyle W_{dummy}\right \|_{2} })^T (\frac{\scriptstyle f^{ALR}}{\left \| \scriptstyle f^{ALR}\right\|_{2}}),y \right )
\end{equation}
Then, we can calculate the loss for the ALI feature: 
\begin{equation}\label{eq9}
	\mathcal{L}^{ALI}_{dummy}=CE\left ((\frac{ \scriptstyle W_{dummy}}{\left\|\scriptstyle W_{dummy}\right \|_{2} })^T (\frac{ \scriptstyle f^{ALI}}{\left \| \scriptstyle f^{ALI}\right\|_{2}}),n+1 \right )
\end{equation}
The total is expressed as:
\begin{equation}\label{eq10}
	\mathcal{L}=	\mathcal{L}_{base} + \lambda 	\mathcal{L}^{ALR}_{dummy} + \beta\mathcal{L}^{ALI}_{dummy}  
\end{equation}
In this equation, the first term ensures base-class performance. The second term facilitates the separation of redundancies from the base-class space, leading to a reduction in intra-base-class space. The third term introduces a dummy class to compress inter-base-class space, thereby creating room for novel classes. The combination of the last two terms alleviates the model's confusion between novel and base classes.

Before the incremental session, we extract base-class average embedding (prototypes) as classifiers for base classes using Eq.\ref{proto}. During the incremental session, we maintain the feature extractor $\Phi(\cdot)$ fixed and employ it to extract features from novel-class examples for inference.

\section{Experiment}
In this section, we introduce benchmark datasets and implementation details, validate our method's effectiveness through qualitative and quantitative analyses, perform ablation experiments to analyze component contributions, and compare our method's performance with state-of-the-art approaches on benchmark datasets.
\subsection{Implementation Details}
\textbf{Datasets \& Split:} Our method is evaluated on CIFAR-100 \cite{krizhevsky2009learning}, \textit{mini}ImageNet \cite{russakovsky2015imagenet}, and CUB200 \cite{wah2011caltech}. CIFAR-100 has 60,000 images (64$\times$64) across 100 classes, \textit{mini}ImageNet has 60,000 images (84$\times$84) representing 100 classes from ImageNet \cite{deng2009imagenet}, and CUB200 comprises 11,788 fine-grained images (224$\times$224). Following FSCIL configuration \cite{tao2020few}, both \textit{mini}ImageNet and CIFAR-100 have 60 base classes initially and 40 novel classes distributed across 8 incremental sessions as 5-way 5-shot tasks. For CUB200, the initial session has 100 base classes, and 100 novel classes are distributed over 10 incremental sessions as 10-way 5-shot tasks. This setup aligns with the experimental settings in \cite{yang2023neural}.

\noindent\textbf{Training Details:} Our training protocol follows \cite{yang2023neural}, utilizing the ResNet-12 for CIFAR-100 and \textit{mini}ImageNet, and pretrained ResNet-18 for CUB200. Batch sizes of 128 and 100 are employed for the base session and incremental sessions, respectively.

\begin{table*}[htbp]
	\centering
	\caption{Top-1 average accuracy on all seen classes during each incremental session on CIFAR-100. Results for the compared methods are sourced from \protect\cite{yang2023neural}.}
 \vspace{-0.3cm}
 \adjustbox{width=16cm}{
	\resizebox {\textwidth} {!} {
		\begin{tabular}{lccccccccccc}
			\toprule
			\multirow{2}{*}{\textbf{Method}} & \multicolumn{9}{c}{\textbf{Accuracy in each session(\%)}} &  \textbf{Average} \\ \cmidrule(l){2-9} 
			& \textbf{0} & \textbf{1} & \textbf{2} & \textbf{3} & \textbf{4} & \textbf{5} & \textbf{6} & \textbf{7} & \textbf{8} & \textbf{Acc.}    \\ \cmidrule(r){1-11}
			CEC \cite{zhang2021few}                              & 73.07      & 68.88      & 65.26      & 61.19      & 58.09      & 55.57      & 53.22      & 51.34      & 49.14 & 59.53         \\
			LIMIT \cite{zhou2022few}                            & 73.81      & 72.09      & 67.87      & 63.89      & 60.70      & 57.77      & 55.67      & 53.52      & 51.23   & 61.84   \\
			Meta FSCIL \cite{chi2022metafscil}                       & 74.50      & 70.10      & 66.84      & 62.77      & 59.48      & 56.52      & 54.36      & 52.56      & 49.97   & 60.79  \\
			Self-promoted \cite{zhu2021self}                       & 64.10      & 65.86      & 61.36      & 57.45      & 53.69      & 50.75      & 48.58      & 45.66      & 43.25   &54.52\\
			FACT \cite{zhou2022forward}                          & 74.60      & 72.09      & 67.56      & 63.52      & 61.38      & 58.36      & 56.28      & 54.24      & 52.10    & 62.24 \\
			Data-free Replay \cite{liu2022few}                 & 74.40      & 70.20      & 66.54      & 62.51      & 59.71      & 56.58      & 54.52      & 52.39      & 50.14    & 60.78      \\
			ALICE \cite{peng2022few}                            & 79.00      & 70.50      & 67.10      & 63.40      & 61.20      & 59.20      & 58.10      & 56.30      & 54.10     & 63.21   \\ 
			Ours                             & \textbf{81.45}     & \textbf{77.02}      & \textbf{72.73}      & \textbf{68.95}      & \textbf{65.75}      & \textbf{63.02}      & \textbf{61.07}      & \textbf{59.01}      & \textbf{56.72}  & \textbf{67.30}   \\\hline
			NC-FSCIL(w/ mixup) \cite{yang2023neural}             & 82.52      & 76.82      & 73.34      & 69.68      & 66.19      & 62.85      & 60.96      & 59.02      & 56.11   & 67.50      \\
			Ours(w/ mixup)                 & \textbf{82.52}      & \textbf{78.00}      & \textbf{73.69}      & \textbf{69.73}      & \textbf{66.27}      &\textbf{63.48}      &\textbf{61.46}      & \textbf{59.66}      & \textbf{57.48}  & \textbf{68.03} \\ \bottomrule
	\end{tabular}}}
	\label{t1}
  \vspace{-0.4cm}
\end{table*}
\subsection{Qualitative and Quantitative Analysis }
This section aims to substantiate key claims regarding our method's efficacy through a series of experiments:
\begin{itemize}
	\item \textbf{Decoupling Redundancy}
	\item \textbf{Compressing Base-Class Space}
	\item \textbf{Alleviating the Collision in Feature Space}
        \item \textbf{Mitigating Base-Novel Confusion}
\end{itemize}
\subsubsection{\textbf{Decoupling Redundancy from Base-class Space}}
\begin{figure}[htbp]
        \vspace{-0.5cm}
	\includegraphics[width=8cm]{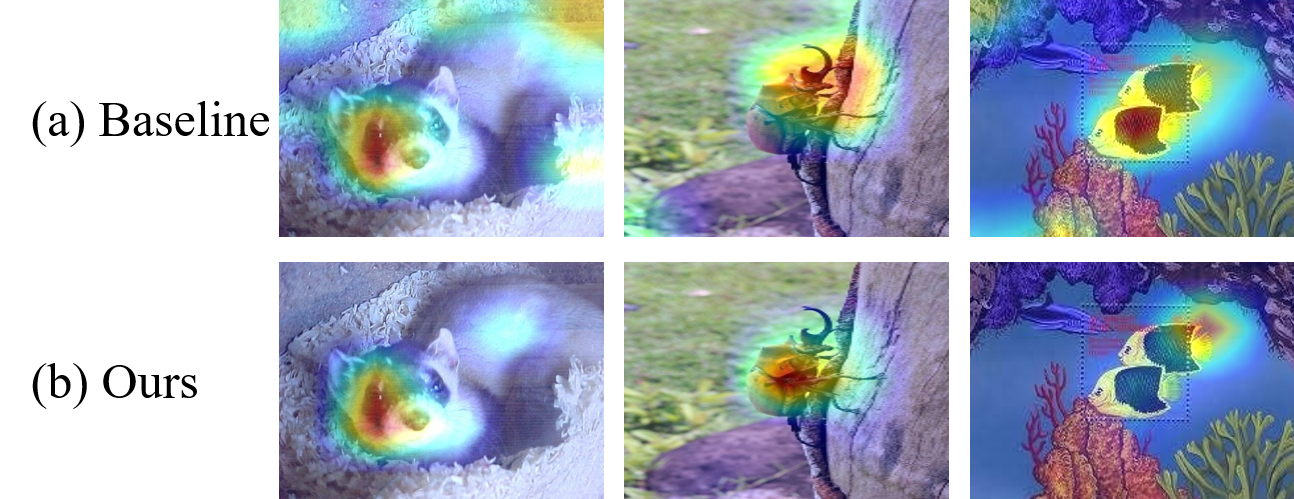}
	\caption{Compared to the baseline method, ours can decouple redundancies and make the model focus on label-relevant regions.}
	\label{v5}
 \vspace{-0.3cm}
\end{figure}

We randomly choose images from the \textit{mini}ImageNet dataset to visualize the feature map. The heat maps in Fig.\ref{v5} show changes in the feature map. The top row represents the baseline, and the bottom row showcases our method. Significantly, our method directs the model's attention to the label-relevant region, effectively decoupling label-irrelevant redundancies from the base-class space.

\subsubsection{\textbf{Compressing Base-class Space}}
\begin{figure}[htbp]
	\centering
	\includegraphics[width=7.5cm]{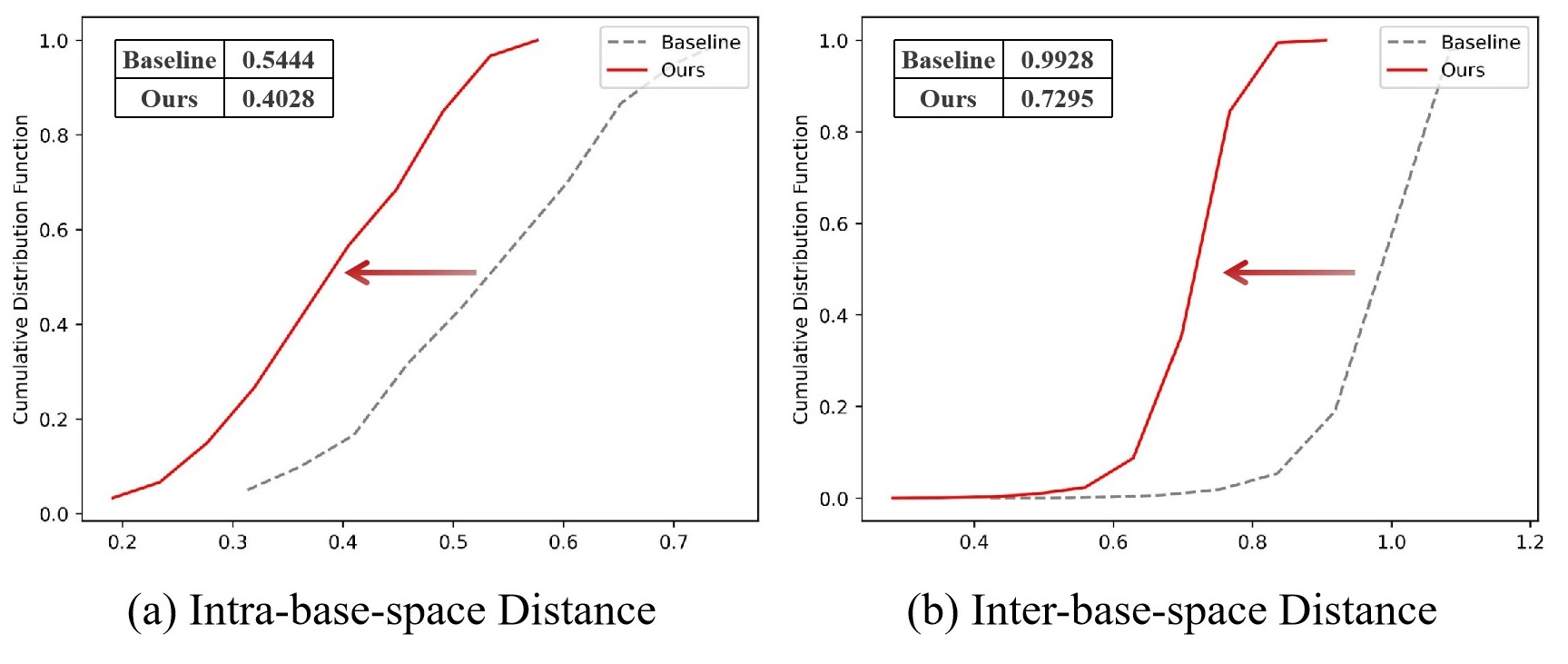}
  \vspace{-0.3cm}
	\caption{The frequencies cumulative distribution diagram depicts inter-base-class and intra-base-class distance, where the abscissa represents cosine distance, and the ordinate shows cumulative frequency. The average cosine distance is presented in the upper left corner. Our method (red solid line) consistently positions to the left of the baseline (gray dotted line), indicating reductions in both inter-base-class and intra-base-class distances.}
 \vspace{-0.4cm}
	\label{vv3}
\end{figure}

To quantitatively showcase our method's efficacy in reducing base-class space, we measure intra-base-class and inter-base-class cosine distances on CIFAR-100 (Fig.\ref{vv3}). Our approach (solid red line) adeptly diminishes both intra-base-class and inter-base-class space, consistently placing cumulative distribution functions to the left of the baseline (gray dashed line).

\subsubsection{\textbf{Alleviating Collisions in Feature Space}}
\begin{figure}[htbp]
\vspace{-0.2cm}
	\centering
	\includegraphics[width=7cm]{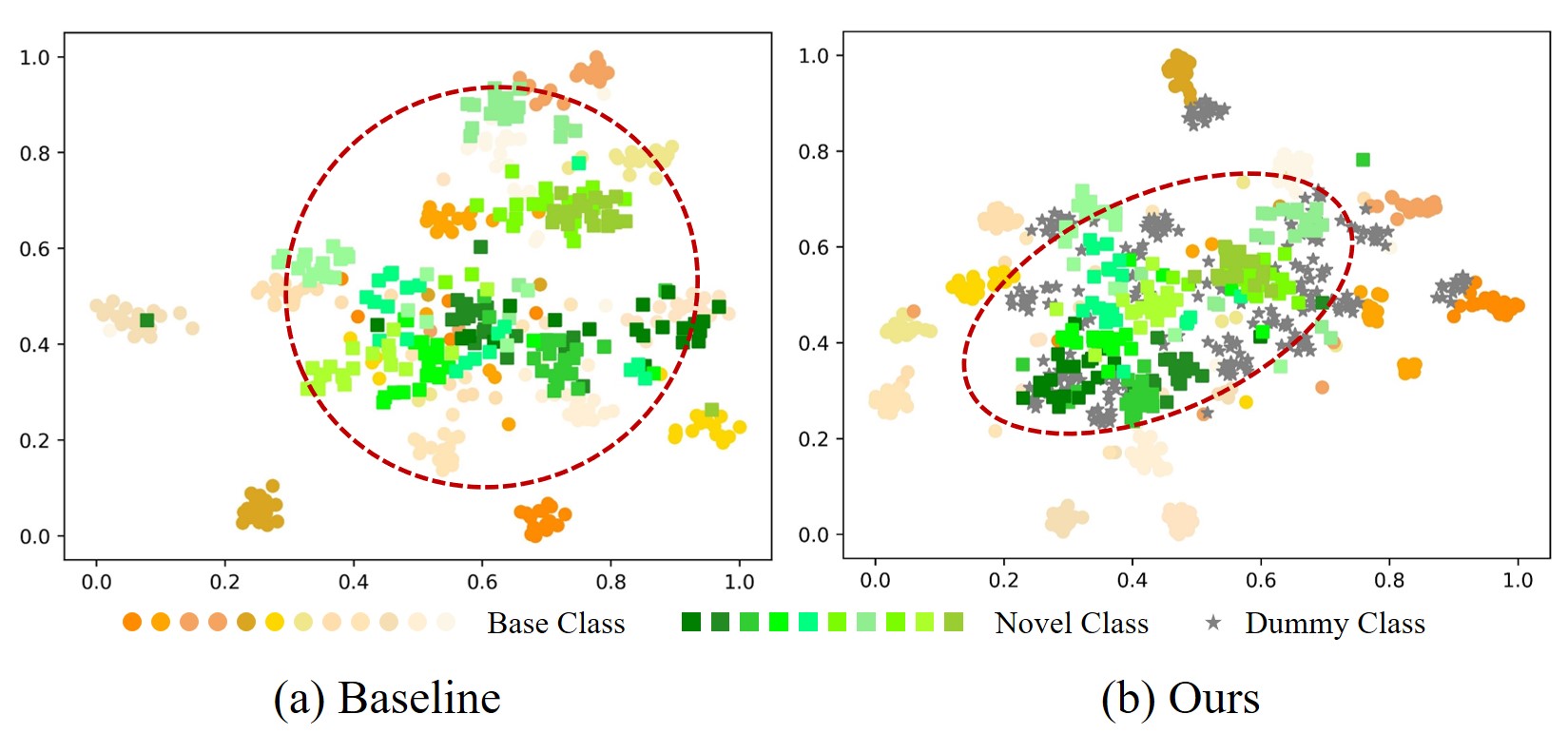}
  \vspace{-0.3cm}
	\caption{The red dotted line outlines the extent of novel classes, showing collisions between novel and base classes in the baseline approach. In contrast, our method introduces a dummy class to reserve space for novel classes, effectively alleviating the collisions.}
	\label{vv1}
 \vspace{-0.4cm}
\end{figure}

To visually highlight our method's efficacy, we randomly selected twelve base and ten novel classes from CIFAR-100, each with twenty samples. Visualizing the feature distribution after inserting novel classes using both baseline and our methods (Fig.\ref{vv1}), the left side shows novel classes (green squares) are mixed with base classes (orange dots) in the baseline. In contrast, on the right side, with our method, novel classes are located within the dummy class (* marked in gray) region and are separated with base classes, showcasing our method's ability to alleviate feature space collisions.
\subsubsection{\textbf{Mitigating the Model's Confusion}}
\begin{figure}[htbp]
 \vspace{-0.2cm}
	\centering
	\includegraphics[width=7cm]{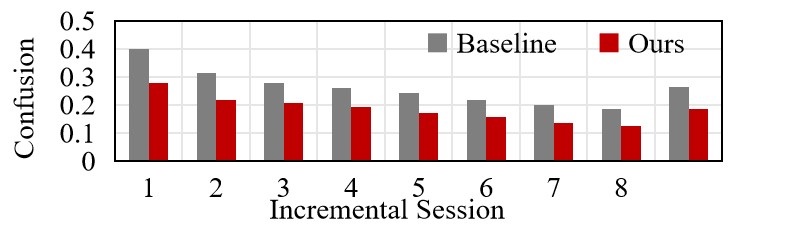}
 \vspace{-0.3cm}
	\caption{Comparing the confusion levels between novel and base classes on the baseline model (gray column) and ours(red column), the results demonstrate that ours effectively alleviates confusion.}
	\label{vv4}
  \vspace{-0.4cm}
\end{figure}

We quantitatively demonstrate our method's effectiveness in reducing the model's confusion between novel and base classes on \textit{mini}ImageNet. Using NN-acc and NA-Acc metrics in Section 2, the larger difference between them indicates increased confusion between base and novel classes. This evaluation, depicted in Fig.~\ref{vv4}, spans both the baseline and our model. Results unequivocally emphasize our approach's success in mitigating the model's confusion.

\subsection{Ablation Study}
We conducted ablation studies on three benchmark datasets, summarizing key findings in Table \ref{t4}. Results highlight our approach's pivotal role in enhancing final session accuracy and improving novel-class generalization. Consistently across benchmark datasets, introducing both $\smash{ \mathcal{L}^{ALR}_{dummy}}$ and $\smash{ \mathcal{L}^{ALI}_{dummy}}$ enhances the model's ability to learn novel classes. The combination of $\smash{\mathcal{L}^{ALI}_{dummy}}$ with $\smash{ \mathcal{L}^{ALR}_{dummy}}$ proves highly effective in reducing confusion, leading to substantial improvements across all classes. These ablation experiments provide insights into contributions of each component, emphasizing the overall effectiveness of our proposed method.

\begin{table}[htbp]
\vspace{-0.2cm}
	\centering
	\caption{Ablation study on three benchmark datasets.}
	\vspace{-0.2cm}
	\adjustbox{width=8cm}{
		\begin{tabular}{l|cc|cc|cc}
			\toprule
			\multicolumn{1}{c}{\multirow{2}{*}{\textbf{Method}}} & \multicolumn{2}{c}{CIFAR-100} & \multicolumn{2}{c}{\textit{mini}ImageNet} & \multicolumn{2}{c}{CUB200}                                          \\ 
			& novel & average      & novel  & average     & novel & average     \\ \cmidrule(r){1-7}
			$\mathcal{L}_{base}$   & 19.62  & 54.53        & 23.50  & 57.64      &40.78 & 58.63      \\
			$\mathcal{L}_{base}+\mathcal{L}^{ALR}_{dummy}$                  & 22.77 & 54.87      & 26.82   & 58.58          & 42.05 & 58.92      \\
			$\mathcal{L}_{base}+\mathcal{L}^{ALI}_{dummy}$                  & 24.40 & 56.28      & 23.75   & 56.94          & 42.59 & 59.61      \\
			$\mathcal{L}_{base}+\mathcal{L}^{ALR}_{dummy}+\mathcal{L}^{ALI}_{dummy}$ 
			& \textbf{26.42}   & \textbf{56.72}     & \textbf{30.52}    & \textbf{59.25}       & \textbf{44.13}    & \textbf{60.20}                   \\ \bottomrule
	\end{tabular}}
	\label{t4}
 \vspace{-0.4cm}
\end{table}

\subsection{Comparison with the State-of-the-Art Methods}
\begin{figure}[htbp]
\vspace{-0.3cm}
	\centering
	\includegraphics[width=8cm]{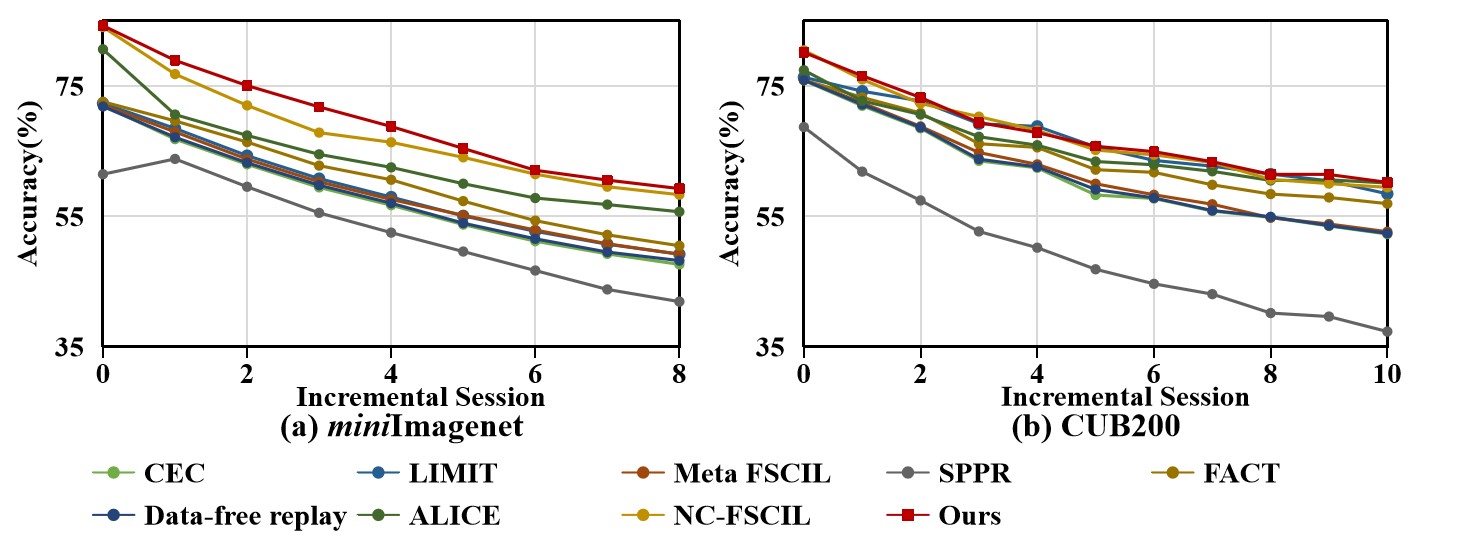}
 \vspace{-0.2cm}
	\caption{Detailed average accuracy during each incremental session on \textit{mini}ImageNet and CUB200.}
	\label{exp}
 \vspace{-0.4cm}
\end{figure}

We comprehensively compare our proposed method with existing approaches across three FSCIL benchmarks. Performance curves for \textit{mini}ImageNet and CUB200 are shown in Figure \ref{exp}, and detailed performance values for CIFAR100 are provided in Table \ref{t1}. Our method consistently outperforms previous approaches in average accuracy across the benchmarks, including a notable improvement over NC-FSCIL \cite{yang2023neural}, a recent top-performing method. Specifically, our method achieves marked improvements of 1.37\%, 0.94\%, and 0.76\% in the last session on CIFAR-100, \textit{mini}ImageNet, and CUB200, respectively. Moreover, we substantiate our superiority by attaining an average accuracy enhancement of 1.74\% on \textit{mini}ImageNet.

\section{Related Work}
Few-Shot Class-Incremental Learning (FSCIL) \shortcite{hou2019learning,rebuffi2017icarl,dong2021few} seamlessly combines Class Incremental Learning (CIL) \shortcite{aljundi2018memory,wu2019large,wang2022learning} and Few-Shot Learning (FSL) \shortcite{chen2019closer,wang2020score,sung2018learning}. FSCIL prioritizes mastering base classes with abundant examples initially and then adeptly addressing novel classes with few instances in each incremental session.

FACT \cite{zhou2022forward} addresses feature distribution discrepancies between novel and base classes by introducing forward compatibility. FACT allocates space for future novel classes through virtual instances from instance fusion, creating a dedicated novel-class space. However, concerns may arise regarding the accuracy of synthesized samples in representing real-world instances.

N-FSCIL \cite{yang2023neural}, the leading approach, pre-assigns classifier prototypes for the entire label spectrum to rectify feature-classifier misalignment for old classes. It reserves space for novel classes by referencing class prototypes to accommodate both base and novel categories. However, this method requires prior knowledge of novel class counts, limiting its practical applicability.

TEEN \cite{wang2023few}, closely related to our work, addresses the lower performance of novel classes in FSCIL, highlighting the model's tendency to misclassify them into base classes. TEEN introduces a training-free prototype calibration method based on semantic similarity between base and novel classes. In contrast, our study thoroughly analyzes the root cause of low novel class performance from a feature distribution perspective, proposing a more effective and widely applicable solution to this issue.
 \vspace{-0.3cm}
\section{Conclusion}
In this paper, we address the challenge of confusion between novel and base classes within the Few-Shot Class-Incremental Learning framework, particularly when using a fixed backbone trained on base classes. Through a thorough analysis of the root cause behind the model's confusion in existing methods – redundancy within the feature space of the base class leading to collisions between novel and base classes in feature space – we introduce the Redundancy Decoupling and Integration (RDI) method. Through comprehensive experiments on benchmark datasets, we empirically validate the efficacy of the RDI method. Our results demonstrate substantial improvements on novel-class generalization and achieve state-of-the-art performance across all classes.

\section*{Acknowledgments}
This work is supported by National Natural Science Foundation of China under grants 62206102, 62376103, 62302184, 62377009, U1936108 and Science and Technology Support Program of Hubei Province under grant 2022BAA046.

\bibliographystyle{named}
\bibliography{final}
\end{document}